%% file: main.tex
\documentclass[conference]{IEEEtran}
\IEEEoverridecommandlockouts
\usepackage{cite}
\usepackage{amsmath,amssymb,amsfonts}
\usepackage{algorithmic}
\usepackage{graphicx}
\usepackage{textcomp}
\usepackage{xcolor}
\usepackage{booktabs} 
\usepackage{multirow}
\usepackage{subfig}
\usepackage{hyperref}
 
\def\BibTeX{{\rm B\kern-.05em{\sc i\kern-.025em b}\kern-.08em
    T\kern-.1667em\lower.7ex\hbox{E}\kern-.125emX}}
\begin{document}

\title{Trajectory Encoding Temporal Graph Networks}
% \author{\IEEEauthorblockN{Anonymous Authors}}
\author{\IEEEauthorblockN{Jiafeng Xiong}
\IEEEauthorblockA{\textit{Department of Computer Science} \\
\textit{University of Manchester}\\
Manchester, United Kingdom \\
jiafeng.xiong@manchester.ac.uk}
\and
\IEEEauthorblockN{Rizos Sakellariou}
\IEEEauthorblockA{\textit{Department of Computer Science} \\
\textit{University of Manchester}\\
Manchester, United Kingdom \\
rizos@manchester.ac.uk}
}

\maketitle

\begin{abstract}
Temporal Graph Networks (TGNs) have demonstrated significant success in dynamic graph tasks such as link prediction and node classification. Both tasks comprise transductive settings, where the model predicts links among known nodes, and in inductive settings, where it generalises learned patterns to previously unseen nodes. Existing TGN designs face a dilemma under these dual scenarios. Anonymous TGNs, which rely solely on temporal and structural information, offer strong inductive generalisation but struggle to distinguish known nodes. In contrast, non-anonymous TGNs leverage node features to excel in transductive tasks yet fail to adapt to new nodes. To address this challenge, we propose Trajectory Encoding TGN (TETGN). Our approach introduces automatically expandable node identifiers (IDs) as learnable temporal positional features and performs message passing over these IDs to capture each node's historical context. By integrating this trajectory-aware module with a standard TGN using multi-head attention, TETGN effectively balances transductive accuracy with inductive generalisation. Experimental results on three real-world datasets show that TETGN significantly outperforms strong baselines on both link prediction and node classification tasks, demonstrating its ability to unify the advantages of anonymous and non-anonymous models for dynamic graph learning \footnote{Code and data are available at: \href{https://github.com/xjiaf/TETGN.git}{https://github.com/xjiaf/TETGN.git}}. 

\end{abstract}


\section{Introduction}
% scarselli_graph_2008,
Graph representation learning and Graph Neural Networks (GNNs)~\cite{bronstein_geometric_2017,wu_comprehensive_2020,liang_survey_2022} have achieved remarkable success in diverse applications by utilising message-passing mechanisms over static graphs~\cite{battaglia_relational_2018}. Nonetheless, many real-world systems such as social networks, communication networks, and recommendation systems~\cite{skarding_foundations_2021,zhang_graph_2021,jiang_graph-based_2022,wu_graph_2022}, are dynamic naturally, where the relationships and entities change over time.

To capture these dynamics, temporal graph representation learning and Temporal Graph Networks (TGNs)~\cite{kumar_predicting_2019,trivedi_dyrep_2019,xu_inductive_2020,rossi_temporal_2020,souza_provably_2022,gravina_long_2024} have emerged as a promising direction, integrating both temporal and structural information into node embeddings. By incorporating time-aware message passing, TGNs adapt to shifting node interactions to predict tasks such as node classification~\cite{xu_adaptive_2019} and link prediction in temporal networks~\cite{ghorbanzadeh_hybrid_2021,tian_knowledge_2022,xue_dynamic_2022}.

In temporal graph representation learning, models typically operate under two main prediction settings: transductive and inductive. Transductive methods predict outcomes for nodes already observed, whereas inductive methods aim to handle previously unseen nodes. Despite notable advancements, developing a TGN that effectively balances transductive and inductive requirements remains challenging. 
This fundamental challenge is reflected in current approaches, where existing methods typically excel in one setting whilst compromising performance in the other. Specifically, temporal graph representation learning can be classified into anonymous and non-anonymous strategies, each demonstrating this inherent trade-off.

Anonymous methods, such as causal anonymous walks (CAW)~\cite{wang_inductive_2022}, rely solely on relative temporal and positional relationships without leveraging explicit node features. This design choice promotes strong inductive performance, as it does not depend on pre-obtained node features or IDs. However, anonymity leads to a shortage in transductive tasks involving complex topologies (e.g. symmetrical or cyclic structures), where the absence of unique node information limits the model's ability to distinguish structurally similar nodes.

By contrast, non-anonymous methods incorporate node features to generate node embedding and make predictions. This inclusion enhances the expressiveness required to solve transductive tasks since node-specific information is unique. Yet, relying on pre-existing features can hinder inductive generalisation for unseen nodes. Moreover, purely feature-based approaches may not adequately capture the networks' relative temporal and positional dynamics.

% To address these issues, we propose \underline{\textbf{T}}rajectory \underline{\textbf{E}}ncoding \underline{\textbf{TGN}}s (\textbf{TETGN}), which seamlessly integrates the strengths of both anonymous and non-anonymous strategies. Our model introduces two key innovations: \emph{automatically extendable learnable node IDs} and a \emph{trajectory encoding} module. The learnable node IDs serve as initial position features whilst being dynamically assignable to unseen nodes, thereby overcoming the inductive limitations of traditional non-anonymous methods. The trajectory encoding module leverages message passing to capture nodes' historical temporal-positional relationships based on these learnable IDs, preserving the generalisation benefits of anonymous approaches whilst maintaining distinctive node-level information crucial for complex transductive tasks.

To address these issues, we propose \underline{\textbf{T}}rajectory \underline{\textbf{E}}ncoding \underline{\textbf{TGN}}s (\textbf{TETGN}s), which integrate the strengths of both anonymous and non-anonymous strategies. Our model introduces two key innovations: \emph{automatically extendable learnable node IDs} and a \emph{trajectory encoding} module. The learnable node IDs serve as initial position features whilst being assignable to unseen nodes, thereby overcoming the inductive limitations of traditional non-anonymous methods. The trajectory encoding module leverages message passing to capture nodes' historical temporal-positional relationships based on these learnable IDs, preserving the generalisation benefits of anonymous approaches whilst maintaining distinctive node-level information crucial for complex transductive tasks.

Architecturally, TETGN extends the MP-TGN framework~\cite{rossi_temporal_2020} by incorporating a parallel \emph{trajectory message passing} module alongside the original encoder. This dual-stream design enables separate processing of node features and trajectory information, with their outputs being stored in memory. Subsequently, a multi-head attention module fuses these complementary representations—node states and trajectory encodings—into final node embeddings. Through this architecture, TETGN achieves better expressiveness whilst improving both transductive and inductive capabilities.

In summary, the main contributions of this paper are:
\begin{itemize}
    \item We propose exploiting\textit{automatically extendable learnable node IDs} as temporal positional features to handle unseen nodes and resolve the transductive indistinguishability issues inherent in anonymous approaches.
    \item We introduce a \emph{trajectory encoding} component via message passing to capture each node’s historical temporal positional relationships, helping non-anonymous methods in inductive generalisation.
    \item We evaluate TETGN through extensive experiments on temporal link prediction and node classification tasks across three real-world benchmark datasets and two masked datasets. Our results demonstrate that TETGN consistently and significantly outperforms strong baselines, achieving superior performance in both transductive and inductive settings. 
\end{itemize}

The remainder of the paper is organised as follows. Section~\ref{sec:related_work} reviews related approaches for dynamic graph learning and position encoding in TGNs. Section~\ref{sec:preliminaries} introduces preliminaries on temporal graphs and formalises the learning problem. Section~\ref{sec:method} details the \textit{trajectory} concept and the TETGN architecture. Section~\ref{sec:experiments} reports experimental settings, performance comparisons, ablation studies, and hyper-parameter analyses. Finally, Section~\ref{sec:conclusion} provides concluding remarks and discusses potential future directions.

\section{Related Work}
\label{sec:related_work}

\subsection{Temporal Graph Neural Networks}
Temporal networks can be modelled as \emph{discrete-time dynamic graphs} (DTDG) or \emph{continuous-time dynamic graphs} (CTDG). In the DTDG setting, a sequence of graph snapshots is recorded at uniform time intervals. Earlier studies~\cite{li_deep_2018,chen_e-lstm-d_2021,goyal_dyngem_2018} apply traditional GNNs~\cite{scarselli_graph_2009} such as GCN~\cite{kipf_semi-supervised_2017} or GAT~\cite{velickovic_graph_2018} to these snapshots and aggregate the resulting embeddings for downstream tasks~\cite{xu_inductive_2020,wang_inductive_2022}. However, these frameworks often assume regular temporal patterns in event arrivals, limiting their ability to effectively model the complex, irregular dynamics inherent in real-world temporal networks.

In contrast, CTDG approaches treat temporal networks as collections of timestamped events with variable inter-event intervals. For instance, DyRep~\cite{trivedi_dyrep_2019} and JODIE~\cite{kumar_predicting_2019} employ RNNs to refresh node embeddings after each event, while TGAT~\cite{xu_inductive_2020}, TGN~\cite{rossi_temporal_2020}, and PINT~\cite{souza_provably_2022} adopt message-passing to aggregate neighbourhood information over time. Meanwhile, random-walk-based methods such as CTDNE~\cite{nguyen_continuous-time_2018} and CAW~\cite{wang_inductive_2022} learn representations by exploring local structures. Furthermore, recent works have explored diverse approaches to enhance dynamic graph representation learning: GraphMixer~\cite{cong_we_2023} utilises efficient neighbour pooling mechanisms, DyGFormer~\cite{yu_towards_2023} adopts specialised Transformer architectures for temporal dynamics, and FreeDyG~\cite{tian_freedyg_2024} introduces innovative node frequency-based modelling.

\subsection{Position Encoding in Temporal Graphs learning}

Position encoding in temporal graphs learning aims to capture structural or temporal information to enhance node embeddings. TGAT~\cite{xu_inductive_2020} extends graph attention networks with sinusoidal time encoding to encode temporal neighbours, whilst CAW~\cite{wang_inductive_2022} and PINT~\cite{souza_provably_2022}  use temporal anonymous walks to count motifs and encode the relative positional information. Despite these advances, methods without node features struggle to discriminate between nodes in complex topological structures, presenting a critical challenge in effective temporal graph learning.

\section{Preliminaries}\label{sec:preliminaries}
\subsection{Notations and Problem Definition} 
In a CTDG temporal graph ${G}(t) = (V(t), E(t), \mathcal{X},\mathcal{E})$ defined over the time domain $T=\{t_1,t_2,\dots\mid t_1<t_2<\dots\}$, $V(t) \subseteq V$ denotes the set of nodes present at timestamp $t$, where $V = \{1, 2, \ldots, n\}$ is the set of all nodes. The set $E(t) \subseteq \{(i, j, t') \mid i, j \in V(t), t' \leq t\}$ represents a sequence of timestamped edge events between nodes $i$ and $j$ at time $t'$. Each event may have an edge feature $e_{ij}(t) \in \mathcal{E}$, and $\mathcal{E} \in \mathbb{R}^k$ is the edge feature set. Each node $i$ has an initial feature ${x}_i \in \mathcal{X}$, where $\mathcal{X} \in \mathbb{R}^d$ is the node feature set. Additionally, the temporal neighbourhood $\mathcal{N}_i(t)$ of node $i$ is defined as $\mathcal{N}_i(t) = \{ (j, t') \mid \exists (j, i, t') \in G(t), t' < t \}$.

In both temporal link prediction and node classification, learning can occur in either transductive or inductive settings~\cite{xiong_survey_2025}. In the transductive setting, the model accesses all nodes during training and predicts relationships among these seen nodes. In contrast, the inductive setting requires the model to generalise its learned temporal patterns to predict interactions involving nodes unseen during training.

\subsection{Message-passing Temporal Graph Networks}

Message-passing Temporal Graph Networks (MP-TGNs)~\cite{rossi_temporal_2020} is a  framework for temporal graph representation learning with three key components: aggregation, update, and memory.

The memory consists of vectors that summarise a node's history and are updated as events occur. This update process involves the \(\text{MEM\_AGG}\) module, which aggregates information from the temporal neighbours of the node \(i\), followed by the \(\text{MEM\_UPDATE}\) module, which typically implements a gated recurrent unit (GRU)~\cite{cho_learning_2014}. When the events involving node \(i\) occur at timestamp \(t\), the event message $m_i(t)$ and node state in memory is updated as:
\begin{equation}
m_i(t) = \text{MEM\_AGG}\left( s_i(t), s_j(t), \Delta t, e_{ij}(t)\right) 
\label{eq:mem_agg}
\end{equation}
\begin{equation}
s_i(t) = \text{MEM\_UPDATE}\left(s_i(t), m_i(t)\right)
\label{eq:mem_update}
\end{equation}
where the initial state \(s_i(0)\) is the initial node feature $x_i$; \(s_i(t)\) denotes the updated state of \(i\) at timestamp \(t\); and $\Delta t$ is the time difference between the current timestamp and the last update timestamp.. 

MP-TGNs compute the node embedding \(z^{(l)}_i(t)\) of node \(i\) at layer \(l\) by recursively applying:
\begin{equation}
\tilde{z}^{(l)}_i(t) = \text{AGG}^{(l)}\bigl(z^{(l-1)}_j(t), \Delta t', e_{ij}(t')\bigr),
\label{eq:agg}
\end{equation}
\begin{equation}
z^{(l)}_i(t) = \text{UPDATE}^{(l)}\left(z^{(l-1)}_i(t), \tilde{z}^{(l)}_i(t)\right)
\label{eq:update}
\end{equation}
where edge features $e_{ij}(t) \in \mathcal{E}$, \(z^{(0)}_i(t) = s_i(t)\) represent the state of \(i\) at timestamp \(t\), $\Delta t'$ is the time difference between the current timestamp and the edge's timestamp $t'$, $(j, t')$ belongs to the temporal neighbourhood $\mathcal{N}_i(t)$ and \(\text{AGG}^{(l)}\) and \(\text{UPDATE}^{(l)}\) are parameterised functions. 

\subsection{Expressiveness and Limitations of TGNs}
\begin{figure}
    \centering
    \includegraphics[width=1\linewidth]{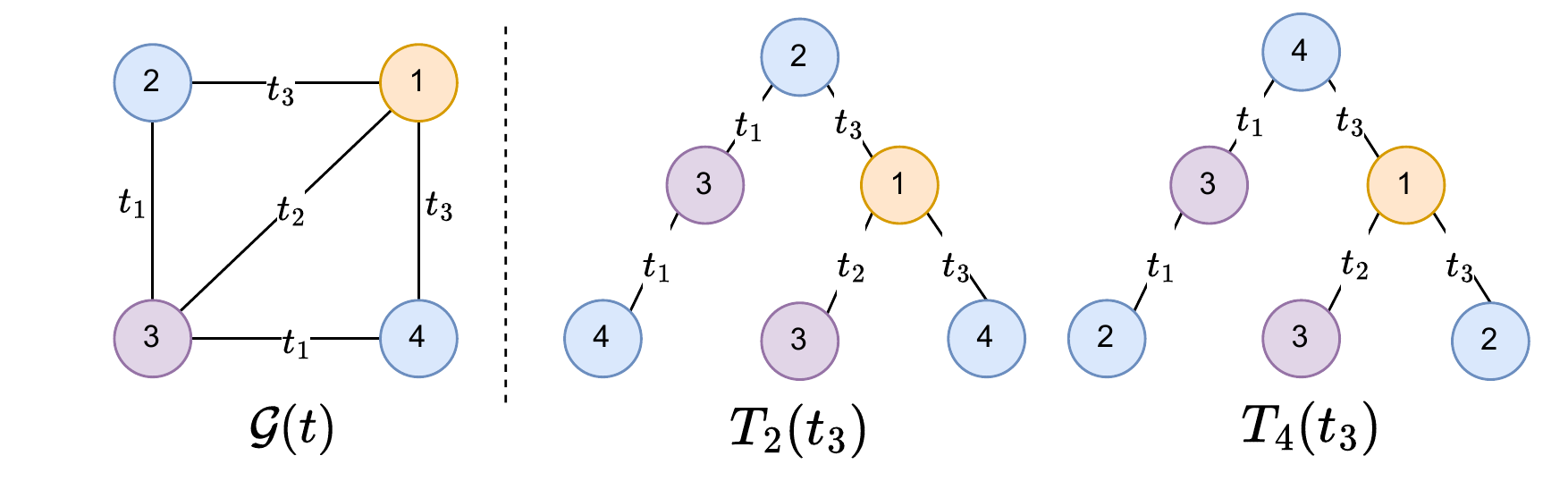}
    \caption{An illustrative example of symmetric and cyclic structures in temporal graphs. The left graph shows a temporal graph where nodes $2$ and $4$ exhibit structural symmetry and are part of a cycle. The right panels display their corresponding temporal computation trees (TCTs), which are isomorphic.}
    \label{fig:mix}
\end{figure}

The Weisfeiler-Lehman (WL) test~\cite{weisfeiler_reduction_1968} serves as a fundamental benchmark for assessing GNN expressiveness, establishing that standard message-passing architectures cannot exceed its discriminative power~\cite{xu_how_2019,morris_weisfeiler_2019}. For temporal graphs, this concept extends to the T-WL test~\cite{souza_provably_2022}, which evaluates a TGN's ability to distinguish nodes in dynamic settings.

Central to this analysis is the \emph{temporal computation tree} (TCT)~\cite{wang_inductive_2022}, denoted as \(T_i(t)\), which captures \(L\) layers of message passing around node \(i\) up to timestamp \(t\). Two nodes \(j\) and \(j'\) cannot be distinguished by an \(L\)-layer TGN if their TCTs, \(T_j(t)\) and \(T_{j'}(t)\), are isomorphic, meaning they share identical structure, timestamps, and embedding.

As illustrated in Figure~\ref{fig:mix}, nodes \(2\) and \(4\) generate isomorphic TCTs in a graph containing both symmetric and cyclic structures. Anonymous TGNs, which rely solely on relative temporal and positional information, fail the T-WL test in such scenarios as they cannot break this structural symmetry without node features.

\section{Methodology}\label{sec:method}
\subsection{Overview}
\begin{figure*}[ht]
    \centering
    \includegraphics[width=1\linewidth]{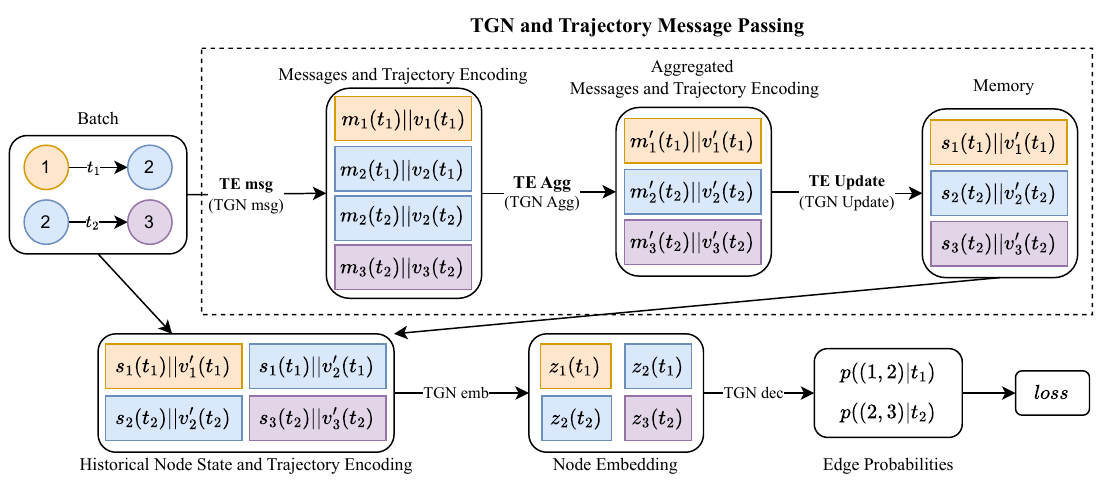}
    \caption{Overview of TETGN architecture. The framework consists of two main components: (a) An encoding phase (top) where temporal interactions (i.e., edges $(1\rightarrow2, 2\rightarrow3)$) are processed through parallel streams - a trajectory message encoding (TE msg) that processes node trajectories and a TGN message encoder that handles standard message passing - with outputs processed through parallel aggregators (TE AGG and TGN AGG) and stored in memory (TE Update and TGN Update); (b) A decoding phase (bottom) where historical node states and trajectory encodings from memory are fused through multi-head attention embedding (TGN emb) to produce the final node embeddings for downstream tasks. Here, $s_i(t_k)$ represents node states in memory, $v_i(t_k)$ and $v'_i(t_k)$ denote raw and aggregated trajectory encodings respectively, $m_i(t_k)$ and $m'_i(t_k)$ represent raw and aggregated messages, and $z_i(t_k)$ denotes the final node embeddings used for downstream tasks.}
    \label{fig:tr_framework}
\end{figure*}

TETGN comprises two main processes—encoding and decoding—as illustrated in Figure~\ref{fig:tr_framework}. The framework extends the MP-TGN architecture \cite{rossi_temporal_2020} by incorporating our trajectory encoding (TE) method.

In the encoding phase (top part of Figure~\ref{fig:tr_framework}), incoming interactions $(i,j,t_k)$ are processed through two parallel streams: the trajectory message encoder and the TGN message computing module, producing trajectory encoding $v_i(t_k)$ and message $m_i(t_k)$, respectively. These outputs are concatenated ($\parallel$) and fed into both trajectory and TGN aggregators as Equation \eqref{eq:mem_agg}, yielding updated representations $v'_i(t_k)$ and $m'_i(t_k)$. Both aggregated values are then stored in the memory module \cite{rossi_temporal_2020}.

The decoding phase (bottom part of Figure~\ref{fig:tr_framework}) focuses on generating node embeddings for downstream tasks. Historical node states and trajectory encodings are retrieved based on previous interactions and concatenated. These combined representations are processed through a multi-head attention module to produce the final node embeddings, which are subsequently used for temporal link prediction. The detailed embedding generation process is presented in Section~\ref{sec:emb}.

\subsection{Trajectory in Message Passing}
\subsubsection{Concept of Trajectory}
We define the trajectory in a temporal graph that captures the evolving history of a node within the network. Specifically, it describes the sequence of interacting node IDs that a node experiences over time. Through a message-passing mechanism, nodes exchange trajectory messages that carry their temporal positional information; each node then updates its trajectory by integrating these incoming messages.

\subsubsection{Automatically Extendable Learnable Node ID}
To leverage the non-anonymous settings, each node is assigned an automatically extendable learnable unique ID vector via a multi-layer perceptron (MLP). This approach helps resolve issues related to symmetric and cyclic structures in temporal networks. The initial ID lookup matrix $\mathbf{M} \in \mathbb{R}^{N \times d}$ maps $N$ nodes to dense vectors of dimension $d$. This node ID serves as the node's initial temporal positional feature $\text{TP}_i(0)$. For a given node $i$, its initial ID vector $\text{ID}_i(0)$ at the beginning is:
\begin{equation}
    \text{TP}_i(0) = \text{ID}_i(0) = \text{MLP}(\mathbf{M}[i])
\end{equation}
where $\mathbf{M}[i]$ denotes the $i$-th row of the lookup matrix $\mathbf{M}$. These temporal positional features will be encoded into trajectory messages through the trajectory message encoding method $\mathrm{TE}(\cdot)$ below, which will be used for message passing.

\subsubsection{Trajectory Messages Encoding}
Each trajectory message encoding $\mathrm{TE}(\cdot)$ may undergo multiple  operations in a nested manner (e.g.\ \(\mathrm{TE}(\mathrm{TE}(\cdot))\)) as it propagates along the TCT based on the message-passing mechanism. To accurately capture both structural (TCT layer-depth) and temporal changes, this encoding must satisfy two key properties:
\begin{enumerate}
    \item As a message traverses deeper layers of the TCT, repeated applications of \(\mathrm{TE}(\cdot)\) should reflect cumulative shifts in \emph{positional} information of layer depth.
    \item When multiple temporal gaps occur sequentially (e.g.\ \(\Delta t_1\) then \(\Delta t_2\)), nested encodings should behave identically to a single encoding of the combined interval \(\Delta t_1 + \Delta t_2\).
\end{enumerate}

Concretely, let \(\mathrm{TE}(x_0, x)\) represent updating a \emph{state} \(x_0\) with two incremental changes \(x\) and $y$. Formally:
\begin{equation}
\mathrm{TE}\bigl(\mathrm{TE}(x_0, x),y\bigr) 
=
\mathrm{TE}\bigl(x_0,x + y\bigr).
\label{eq:te_two_step}
\end{equation}
Suppose we define
\begin{equation}
\mathrm{TE}(x_0, x) 
=
x_0 \cdot f(x),
\end{equation}
where \(f(\cdot)\) encodes the partial update. Then,
\begin{equation}
    \mathrm{TE}\bigl(\mathrm{TE}(x_0, x),y\bigr) 
=
x_0 \cdot f(x)\cdot f(y),
\end{equation}
and
\begin{equation}
    \mathrm{TE}\bigl(x_0,x + y\bigr) 
=
x_0 \cdot f\bigl(x + y\bigr).
\end{equation}
Hence, according to the time consistency required for the second property:
\begin{equation}
f(x)\cdot f(y) 
=
f\bigl(x + y\bigr),
\label{eq:temporal_cauchy}
\end{equation}
which implies that $f(\cdot)$ is a \emph{Cauchy function}. Consequently, we adopt a generic exponential function belonging to the Cauchy function family to represent time decay~\cite{wen_trend_2022}:
\begin{equation}
\text{TE}_{\mathrm{exp}}(x,\Delta t) 
=
\alpha  x  \exp({-\beta \Delta t}),
\label{eq:exp_encoder}
\end{equation}
where \(x\) is the node temporal positional feature, \(\alpha\) is a step-count hyperparameter, \(\beta\) the time decay factor, and \(\Delta t\) the time interval between updates. The power of \(\alpha\) naturally increases with the layer depth as the trajectory message passes through the TCT, reflecting the positional information of the node trajectory, whilst \(\exp({-\beta\Delta t})\) ensures coherent temporal accumulation. For instance, if a trajectory message \(\alpha x_0 \exp({-\beta\Delta t_1})\) is encoded again at a subsequent layer with an additional interval \(\Delta t_2\), then
\begin{align}
\alpha \bigl(\alpha x_0 &\exp({-\beta\Delta t_1})\bigr)\exp({-\beta\Delta t_2}) \nonumber \\
&=
\alpha^2 x_0\exp({-\beta(\Delta t_1 + \Delta t_2)}),
\end{align}
where the power of \(\alpha\) increases by 1 to represent advancing one layer in the TCT, and \(\Delta t_1+\Delta t_2\) tracks the total time elapsed. Thus, this design preserves a coherent representation of both layer-depth progression and accumulated time.

\subsubsection{Trajectory Aggregation}
Similar to standard message passing, node \(i\) at timestamp \(t\) aggregates trajectory messages from its temporal neighbours \(\mathcal{N}_i(t)\). Each neighbour \(j\) provides its latest trajectory message encoding \(v_j(t')\), which is combined with the node’s temporal positional feature \(v_i(t)\) using an associative aggregator (e.g.\ summation):
\begin{equation}
    v'_i(t)=\text{TE\_AGG}(v_i(t), v_j(t')) = v_i(t) + \sum_{(j, t') \in \mathcal{N}_i(t)} v_j(t'),
\end{equation}
where $\text{TE\_AGG}$ is the trajectory aggregation function and node $j$ is the temporal neighbours of node $i$. This approach implicitly tracks higher-order connections within the TCT without explicitly enumerating them. 

\subsubsection{Temporal Positional Feature Update}
The Trajectory Message Passing module will update and derive a new temporal positional feature for the node $i$ based on this learned node ID and the trajectory messages from neighbouring nodes once there is an interaction of node $i$. 

We define the updated temporal positional feature, $\text{TP}_i(t)$, at the timestamp $t$ of the node $i$ based on the learned current ID $\text{ID}_i(t)$ and the aggregated trajectory message  $v'_i(t)$ of the node $i$ just at timestamp $t$:
\begin{equation}
    \text{TP}_i(t)  = \text{TE\_UPDATE}(\text{ID}_i(t), v'_i(t)) = \text{ID}_i(t) + v'_i(t)
\end{equation}
where $\text{TE\_UPDATE}$ is the update function of trajectory encoding, $\text{ID}_i(t)$ is the newly learned ID at current timestamp $t$. The updated new temporal positional feature \(\text{TP}_i(t)\) is stored in memory for future message passing, similar to the memory module in MP-TGN~\cite{rossi_temporal_2020}.

\subsubsection{Complexity Implications}
The TGN has time complexity $\mathcal{O}\left(L(\bar{n} + d)d |E|\right)$, where $\bar{n}$ is the average node degree, $L$ is the number of layers, $|E|$ is the number of edges and \(d\) the embedding dimension. Since TETGN has an extra trajectory message-passing module, time complexity becomes $\mathcal{O}\left(2L(\bar{n} + d)d |E|\right)$. Besides, the storage complexity remains $\mathcal{O}(2|V|d)$. Despite the extra trajectory message-passing module, TETGN's computational overhead is merely a constant factor which is asymptotically negligible in Big-O notation, making it a highly efficient enhancement to TGN.

\subsection{Node Embedding and Prediction}\label{sec:emb}

Following MP-TGN~\cite{rossi_temporal_2020}, TETGN maintains historical states \(s_i(t_k)\) alongside trajectory encodings \(v_i'(t_k)\). These are concatenated and passed through MLPs, then further fused by multi-head attention layers~\cite{vaswani_attention_2017}  to get final embedding \(z_i(t_k)\) for downstream tasks:
\begin{equation}
z_i(t_k) 
=\text{Multi-head\_Attn}(
\text{MLP}([s_i(t_k) 
\parallel
\text{MLP}'(v_i'(t_k))])).
\end{equation}

\section{EXPERIMENTS}\label{sec:experiments}
\subsection{Experiment Setup}
\subsubsection{Datasets}
\begin{table}[ht]
\centering
\caption{Summary of the datasets.}
\label{tab:summary_statistics}

\begin{tabular}{@{}lccc}
\toprule
\textbf{Dataset} & \textbf{\#Nodes} & \textbf{\#Events} & \textbf{\#Edge feat.} \\ \midrule
Reddit  & 10984& 672,447 & 172 \\
Wikipedia & 9,227& 157,474 & 172 \\
Reddit-fm& 10,984& 672,447& -   \\
Wikipedia-fm& 9,227& 157,474& -   \\
LastFM  & 1,980& 1,293,103 & -   \\
\bottomrule
\end{tabular}
\label{tab:datasets}
\end{table}

To evaluate the performance of TETGN, this study applies the methodology to temporal link prediction on three real-world datasets, Wikipedia, Reddit, LastFM and two masked versions Wikipedia-fm, and Reddit-fm \cite{kumar_predicting_2019}, closely adhering to the protocols established by \cite{xu_inductive_2020, rossi_temporal_2020}. The dataset statistics are summarised in Table~\ref{tab:datasets}. For Wikipedia and Reddit, edge feature vectors are intact, while for Wikipedia-fm and Reddit-fm, these edge features are masked to assess the performances with obscured data. LastFM, however, is a completed non-attributed network. For consistency with earlier studies, all datasets employ zero vectors for node features. 

\subsubsection{Baselines}

\begin{table*}[htbp]
\centering
\caption{Comparison of temporal link prediction performance (average precision, mean$\pm$std, over 10 runs) by different models. The best results are marked in bold, and the best baseline results are underlined.}
\label{tab:model_performance}
\begin{tabular}{@{}lcccccc@{}}
\toprule
\textbf{Setting} & \textbf{Model} & \textbf{Reddit} & \textbf{Wikipedia} & \textbf{Reddit-fm}& \textbf{Wikipedia-fm}& \textbf{LastFM} \\ \midrule
\multirow{9}{*}{\rotatebox[origin=c]{90}{Transductive}} 
& GAT             & 97.33 $\pm$ 0.2 & 94.73 $\pm$ 0.2 & -              & -            & -              \\
& GraphSAGE       & 97.65 $\pm$ 0.2 & 93.56 $\pm$ 0.3 & -              & -            & -              \\
\cmidrule(lr){2-7}
& Jodie           & 97.11 $\pm$ 0.3 & 94.62 $\pm$ 0.5 & 96.94 $\pm$ 0.8& 95.19 $\pm$ 0.5& 69.32 $\pm$ 1.0 \\
& DyRep           & 97.98 $\pm$ 0.1 & 94.59 $\pm$ 0.2 & 96.95 $\pm$ 1.0& 94.58 $\pm$ 0.2& 69.24 $\pm$ 1.4 \\
& TGAT            & 98.12 $\pm$ 0.2 & 95.34 $\pm$ 0.1 & 75.11 $\pm$ 0.1& 77.91 $\pm$ 0.5& 54.77 $\pm$ 0.4 \\
& MP-TGN          & 98.70 $\pm$ 0.1 & 98.46 $\pm$ 0.1 & \underline{98.54 $\pm$ 0.1}& 98.54 $\pm$ 0.1& 80.69 $\pm$ 0.2 \\
& CAW-N           & 98.39 $\pm$ 0.1 & \underline{98.63 $\pm$ 0.1}& 98.41 $\pm$ 0.1& \underline{98.58 $\pm$ 0.1} & \underline{81.29 $\pm$ 0.1}\\
& PINT            &  \underline{98.79 $\pm$ 0.1}& 98.40 $\pm$ 0.1& 97.35 $\pm$ 0.1& 98.55 $\pm$ 0.1& 77.06 $\pm$ 1.5\\
\cmidrule(lr){2-7}
& TETGN (ours)     & \textbf{98.90 $\pm$ 0.1}& \textbf{98.74 $\pm$ 0.1} & \textbf{98.79 $\pm$ 0.1}& \textbf{98.73 $\pm$ 0.1} & \textbf{83.26 $\pm$ 0.1} \\
\midrule
\multirow{9}{*}{\rotatebox[origin=c]{90}{Inductive}} 
& GAT             & 95.37 $\pm$ 0.3 & 91.27 $\pm$ 0.4 & -              & -            & -              \\
& GraphSAGE       & 96.27 $\pm$ 0.2 & 91.09 $\pm$ 0.3 & -              & -            & -              \\
\cmidrule(lr){2-7}
& Jodie           & 94.36 $\pm$ 1.1 & 93.11 $\pm$ 0.4 & 93.77 $\pm$ 1.7& 93.77 $\pm$ 0.3& 80.32 $\pm$ 1.4 \\
& DyRep           & 95.68 $\pm$ 0.2 & 92.05 $\pm$ 0.3 & 93.33 $\pm$ 0.2& 92.64 $\pm$ 0.5& 82.03 $\pm$ 0.6 \\
& TGAT            & 96.62 $\pm$ 0.3 & 93.99 $\pm$ 0.3 & 69.72 $\pm$ 0.1& 78.45 $\pm$ 0.5& 56.76 $\pm$ 0.9 \\
& MP-TGN          & 97.55 $\pm$ 0.1 & 97.81 $\pm$ 0.1 & 97.26 $\pm$ 0.1& 97.80 $\pm$ 0.2& 84.66 $\pm$ 0.1 \\
& CAW-N           & 97.81 $\pm$ 0.1 & \underline{98.22 $\pm$ 0.1}& \underline{97.45 $\pm$ 0.1}& \underline{98.15  $\pm$ 0.1}& \underline{85.67 $\pm$ 0.5}\\
& PINT            & \underline{98.25 $\pm$ 0.1}& 97.14  $\pm$ 0.1& 94.72 $\pm$ 0.1& 97.42 $\pm$ 0.2 & 81.57 $\pm$ 0.1\\
\cmidrule(lr){2-7}
& TETGN  (ours)    & \textbf{98.29 $\pm$ 0.1}&  \textbf{98.28 $\pm$ 0.1} & \textbf{98.26 $\pm$ 0.2}& \textbf{98.23 $\pm$ 0.2} & \textbf{88.11 $\pm$ 0.1} \\
\bottomrule
\end{tabular}
\end{table*}

We compare TETGN against both static and temporal GNN baselines. \textbf{GAT}~\cite{velickovic_graph_2018} and \textbf{GraphSAGE}~\cite{hamilton_inductive_2017} represent popular static architectures extended to dynamic tasks by ignoring event timestamps. In contrast, \textbf{Jodie}~\cite{kumar_predicting_2019}, \textbf{DyRep}~\cite{trivedi_dyrep_2019}, \textbf{TGAT}~\cite{xu_inductive_2020}, and \textbf{MP-TGN}~\cite{rossi_temporal_2020} are non-anonymous TGNs. Meanwhile, \textbf{CAW-N}~\cite{wang_inductive_2022} and \textbf{PINT}~\cite{souza_provably_2022} are anonymous models. The results were obtained using the implementations and guidelines from the official repositories. 

Since the experiments adopt a similar setup as MP-TGN~\cite{rossi_temporal_2020,souza_provably_2022}, their reported baseline numbers are used for classic non-anonymous TGNs including GAT, GraphSAGE, Jodie, DyRep, TGAT, and MP-TGN in Wikipedia, Reddit, and LastFM. The remaining results were obtained using the implementations and guidelines available from the official repositories. For the masked dataset Wikipedia-fm and Reddit-fm, the same settings of Wikipedia and Reddit are adopted respectively. This selection of baselines allows us to assess TETGN’s ability to handle both static and dynamic contexts, under anonymous or non-anonymous assumptions.

\subsubsection{Implementation Details}
The experiments use a 70\%-15\%-15\% (train-val-test) temporal split for all datasets, with average precision (AP) as the performance metric. Results are analysed for transductive (seen nodes) and inductive (new nodes) predictions, reporting the mean and standard deviation of AP over 10 runs. All methods are implemented using PyTorch, accelerated by an NVIDIA V100 GPU. The general settings include a batch size of $200$ and $5$ negative samples. For TETGN, the trajectory encoding dimension $d$ is tuned in $\{4,12,20,28\}$, $\beta$ in $\{0.1,1\}$, $\alpha$ in $\{1,2\}$, and the learning rate in $[1e-6, 1e-4]$. Other settings are the same as MP-TGNs.

\subsection{Performance Comparison}
Table~\ref{tab:model_performance} presents the results with the following observations:
\begin{itemize}
    \item Comparing the performance of baselines and TETGN on the Wikipedia, Wikipedia-fm, Reddit, and Reddit-fm datasets shows that TETGN exhibits the smallest performance drop among all methods. This indicates TETGN's low dependency on edge features, highlighting its robustness in capturing and leveraging the inherent temporal dynamics of the graph structure.
    \item On the non‐attributed LastFM dataset, TETGN significantly outperforms the baselines, demonstrating its effectiveness when node and edge features are absent. This underscores TETGN’s superior ability to capture and express temporal graph structural information, with improvements of 2.4\% and 2.8\% in the transductive and inductive settings, respectively.
    \item When comparing baselines from non-anonymous settings such as MP-TGN with those from anonymous settings like CAW-N, CAW-N generally outperforms the non-anonymous baselines, except on the Reddit-FM dataset. This underscores the importance of capturing relative positional relationships. Moreover, TETGN performs even better, leveraging the advantages of the non-anonymous setting for learning network attributes while also effectively capturing relative positional relationships.
\end{itemize}
Overall, TETGN consistently performs better than all baselines on five datasets with both inductive and transductive AP.

\begin{table*}[ht]
\centering
\caption{Ablation study of trajectory message encoder. ``(rand feat.)" means with randomly initialised node features and ``(\textit{w/o.} exp.)" means with a raw ID from the lookup table rather than the exponential encoding.}
\label{tab:ablation}
\begin{tabular}{@{}lcccccc@{}}
\toprule
\textbf{Setting} & \textbf{Model} & \textbf{Reddit} & \textbf{Wikipedia} & \textbf{Reddit-fm}& \textbf{Wikipedia-fm}& \textbf{LastFM} \\ \midrule
\multirow{3}{*}{Transductive} 
& MP-TGN          & 98.70 $\pm$ 0.1 & 98.46 $\pm$ 0.1 & 98.54 $\pm$ 0.1& 98.40 $\pm$ 1.4& 80.69 $\pm$ 0.2 \\
& MP-TGN (rand feat.)   & 98.71 $\pm$ 0.1& 98.49 $\pm$ 0.2& 98.50 $\pm$ 0.2& 98.39 $\pm$ 0.3& 74.71 $\pm$ 1.5\\
& TETGN (\textit{w/o.} exp.)& 98.83 $\pm$ 0.3& 98.68 $\pm$ 0.2&  98.56 $\pm$ 0.1& 98.59 $\pm$ 0.2& 82.52 $\pm$ 0.2\\
& TETGN& \textbf{98.90 $\pm$ 0.2}& \textbf{98.74 $\pm$ 0.1}& \textbf{98.79 $\pm$ 0.1}& \textbf{98.73 $\pm$ 0.1}& \textbf{83.26 $\pm$ 0.1} \\
\midrule
\multirow{3}{*}{Inductive} 
& MP-TGN             & 97.55 $\pm$ 0.1 & 97.81 $\pm$ 0.1 & 97.26 $\pm$ 0.1& 97.80 $\pm$ 0.2& 84.66 $\pm$ 0.1 \\
& MP-TGN (rand feat.)   & 97.41 $\pm$ 0.4& 97.77 $\pm$ 0.3& 97.16 $\pm$ 0.2& 97.67 $\pm$ 0.1& 79.20 $\pm$ 3.9\\
& TETGN (\textit{w/o.} exp.)& 97.93 $\pm$ 0.2& 98.15 $\pm$ 0.1& 97.31 $\pm$ 0.2& 98.12 $\pm$ 0.1& 87.51 $\pm$ 0.1\\
& TETGN& \textbf{98.29 $\pm$ 0.1}& \textbf{98.28 $\pm$ 0.1}& \textbf{98.26 $\pm$ 0.1}& \textbf{98.23 $\pm$ 0.2}& \textbf{88.11 $\pm$ 0.1} \\
\bottomrule
\end{tabular}
\end{table*}

\subsection{Ablation Study}
\subsubsection{Ablation study on a non-anonymous setting}
To validate the effect of the learnable ID, experiments were conducted using a version of MP-TGN with randomly initialised node features, referred to as MP-TGN (rand feat.), and a variant of TETGN where the exponential trajectory message encoder was removed, retaining only the learnable ID module, referred to as TETGN (\textit{w/o.} exp.). These models were trained using the same settings that were used for the results in Table~\ref{tab:model_performance}. The results are reported in Table~\ref{tab:ablation}. The following observations are made:
\begin{itemize}
    \item Comparing MP-TGN (rand feat.) with MP-TGN shows little change in performance on the attributed datasets Wikipedia and Reddit, but a decline on the three non-attributed datasets. This suggests that using node IDs to generate a lookup table for IDs offers minimal benefit to TGNs and can even harm performance in non-attributed datasets where capturing the graph structure is crucial.
    \item When comparing MP-TGN (rand feat.) with TETGN (\textit{w/o.} exp.), TETGN (\textit{w/o.} exp.) demonstrates a significant performance improvement. This indicates that the learnable ID, even without the Exponential Trajectory Message Encoder, provides a meaningful enhancement by effectively capturing node-specific information. This enhancement suggests that the non-anonymous setting plays a crucial role in improving the model's ability to represent nodes in a way that better captures the temporal and structural nuances of the graph.
    \item When comparing the results of the MP-TGN and TETGN (\textit{w/o.} exp.) experiments with those of the MP-TGN (rand feat.) and TETGN (\textit{w/o.} exp.) experiments, it is found that the independent trajectory encoding architecture, which captures temporal and relative positional relationships, performs better than simply using message-passing to represent node features. This suggests that the trajectory encoding framework may offer advantages in modelling the temporal dynamics of temporal graphs.
\end{itemize}

\subsubsection{Ablation study on exponential trajectory message encoder}
Further investigation into how the trajectory message encoder contributes to the overall performance of TETGN reveals that TETGN consistently outperforms TETGN (\textit{w/o.} exp.). This suggests that the exponential trajectory message encoder is critical for the model. The consistent performance improvement highlights the importance of this component in leveraging the temporal and structural information, thereby enhancing the overall effectiveness of TETGN.

\subsection{Hyper-parameter Analysis on \texorpdfstring{$\alpha$ and $\beta$}{alpha and beta}}
\begin{figure}[htbp]
    \centering    
    \subfloat[$\alpha$ on Wikipedia]
    {
        \begin{minipage}[t]{0.23\textwidth}
            \centering    
            \includegraphics[width=1\textwidth]{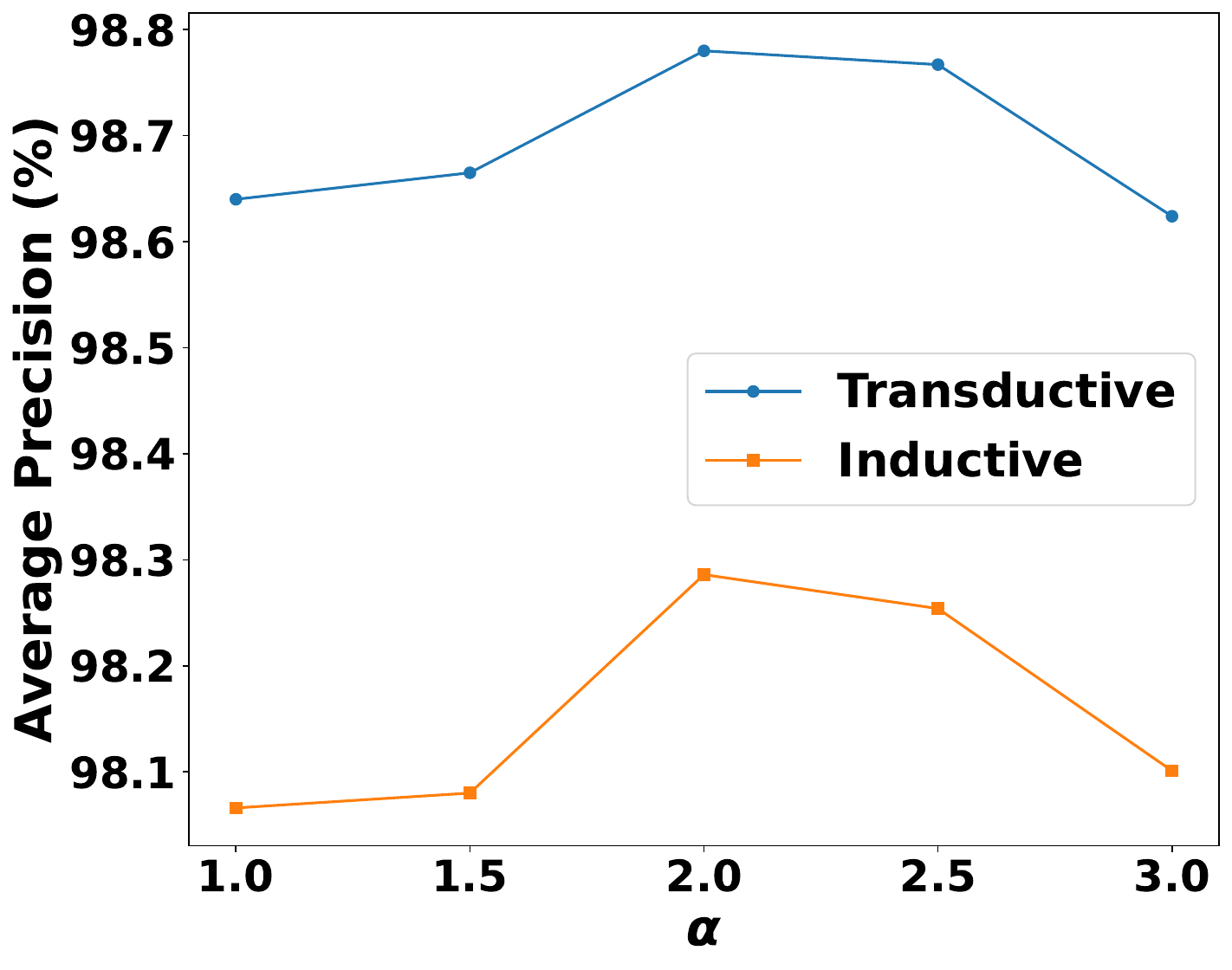}           \end{minipage}
    }
    \subfloat[$\beta$ on Wikipedia] 
    {
        \begin{minipage}[t]{0.23\textwidth}
            \centering    
            \includegraphics[width=1\textwidth]{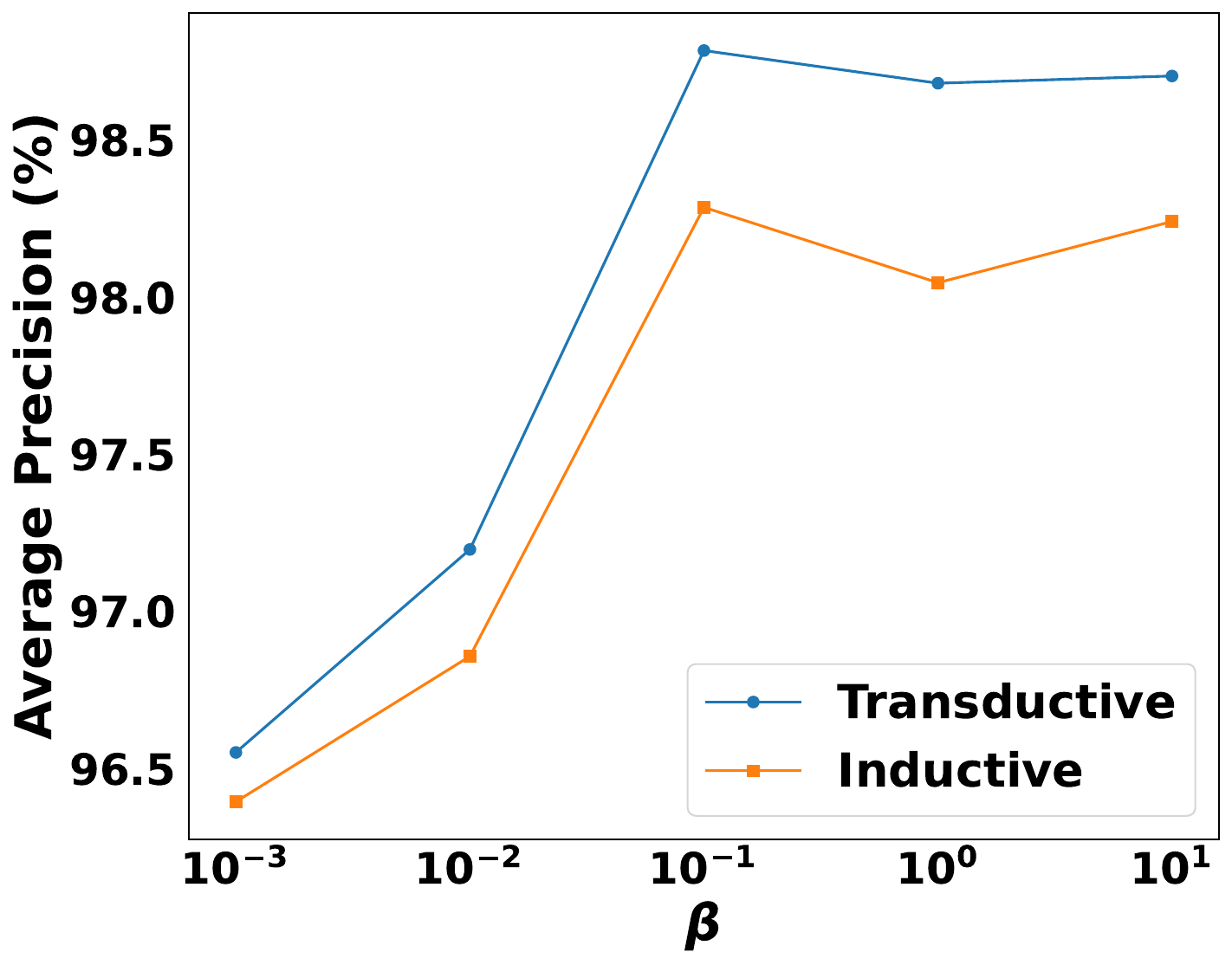}           \end{minipage}
    }
    \caption{Hyper-parameter analysis on \(\alpha\) and $\beta$.} 
    \label{fig:alpha_beta}  
\end{figure}
Figure~\ref{fig:alpha_beta} illustrates the impact of the hyper-parameters including scale parameter \(\alpha\) and time decay factor \(\beta\) on performance. Two experiments were conducted: one with \( \alpha \) fixed at 2 while \( \beta \) was varied, and another with \( \beta \) fixed at 0.1 while \( \alpha \) was varied. The mean AP for each setting, averaged over 10 runs, is reported. 

In both cases, the performance exhibits a hump-shaped curve. Optimal performance is achieved with \( \alpha \) around 2 and \( \beta \) near 0.1. Values deviating significantly from these optima result in lower accuracy, highlighting the sensitivity of the system to the choice of these parameters. Notably, setting either \( \alpha \) or \( \beta \) to 1 leads to a marked performance decline; this suggests that a value of 1 effectively renders the corresponding parameter inactive, thereby diminishing its contribution.

\subsection{Further Hyper-parameter Analysis on \texorpdfstring{$\alpha$, $\beta$ and $d$}{alpha, beta and d}}

\begin{figure}[htbp]
    \centering    
    \subfloat[Inductive on Wikipedia]
    {
        \begin{minipage}[t]{0.23\textwidth}
            \centering    
            \includegraphics[width=1\textwidth]{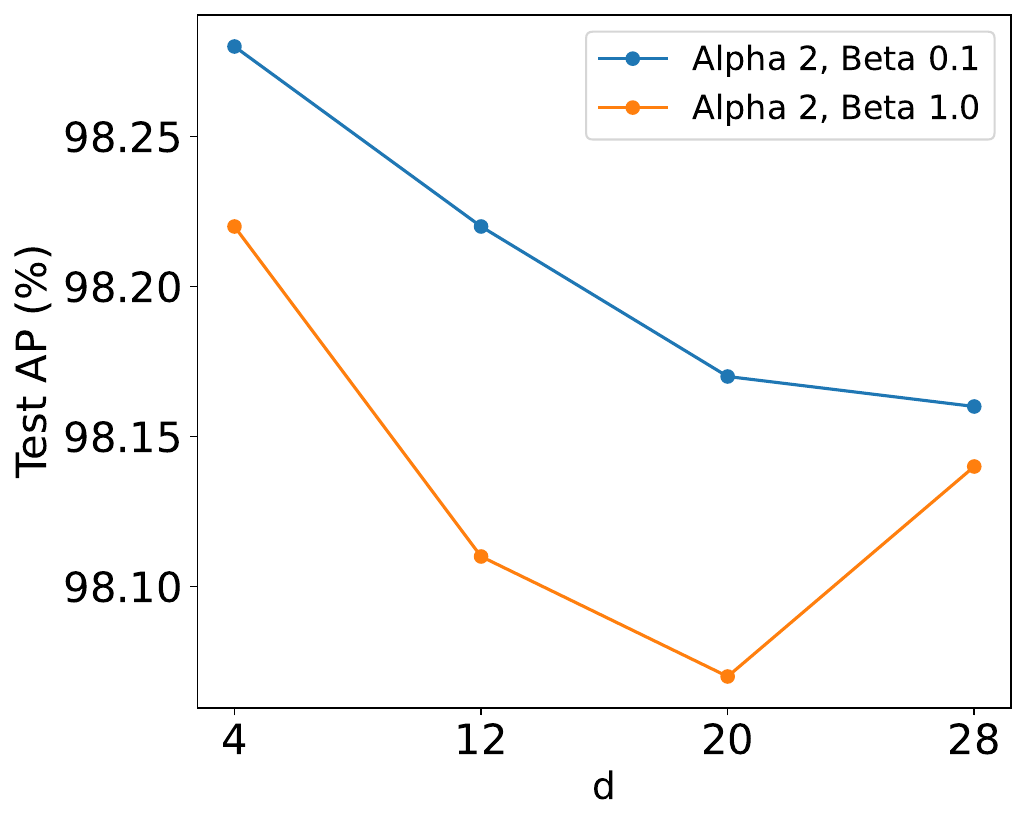}   
        \end{minipage}
    }
    \subfloat[Transductive on Wikipedia] 
    {
        \begin{minipage}[t]{0.23\textwidth}
            \centering    
            \includegraphics[width=1\textwidth]{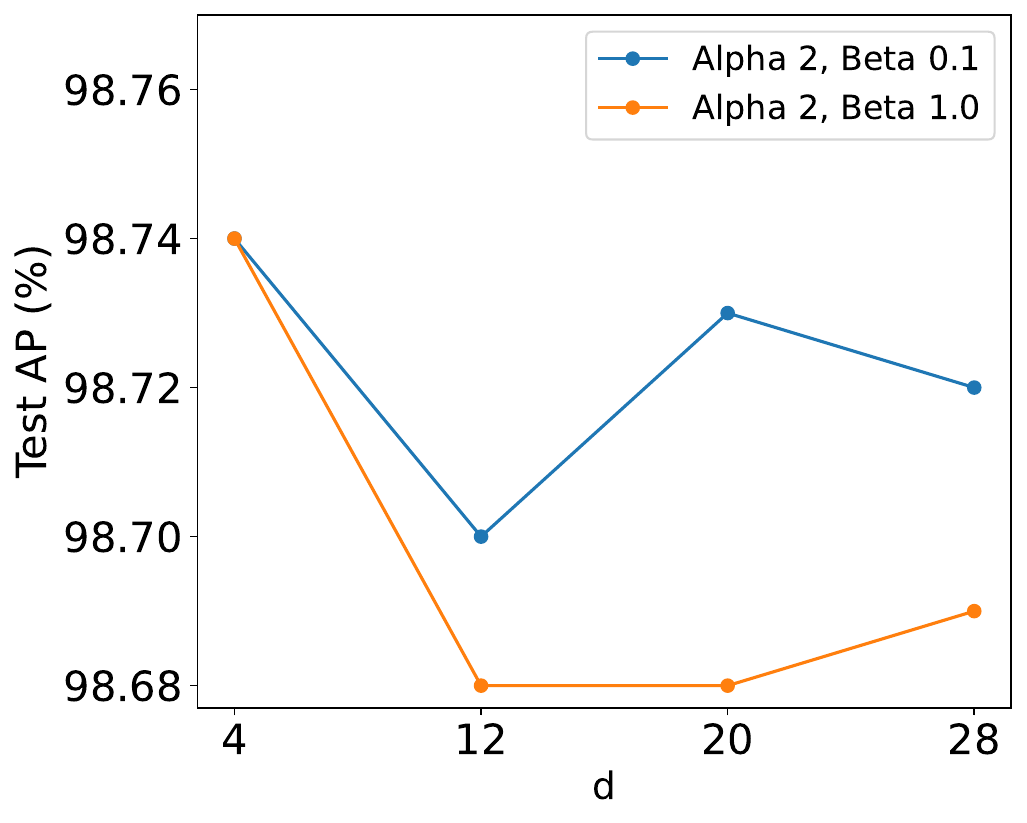}   
        \end{minipage}
    }
    
    \subfloat[Inductive on Wikipedia-fm] 
    {
        \begin{minipage}[t]{0.23\textwidth}
            \centering      
            \includegraphics[width=1\textwidth]{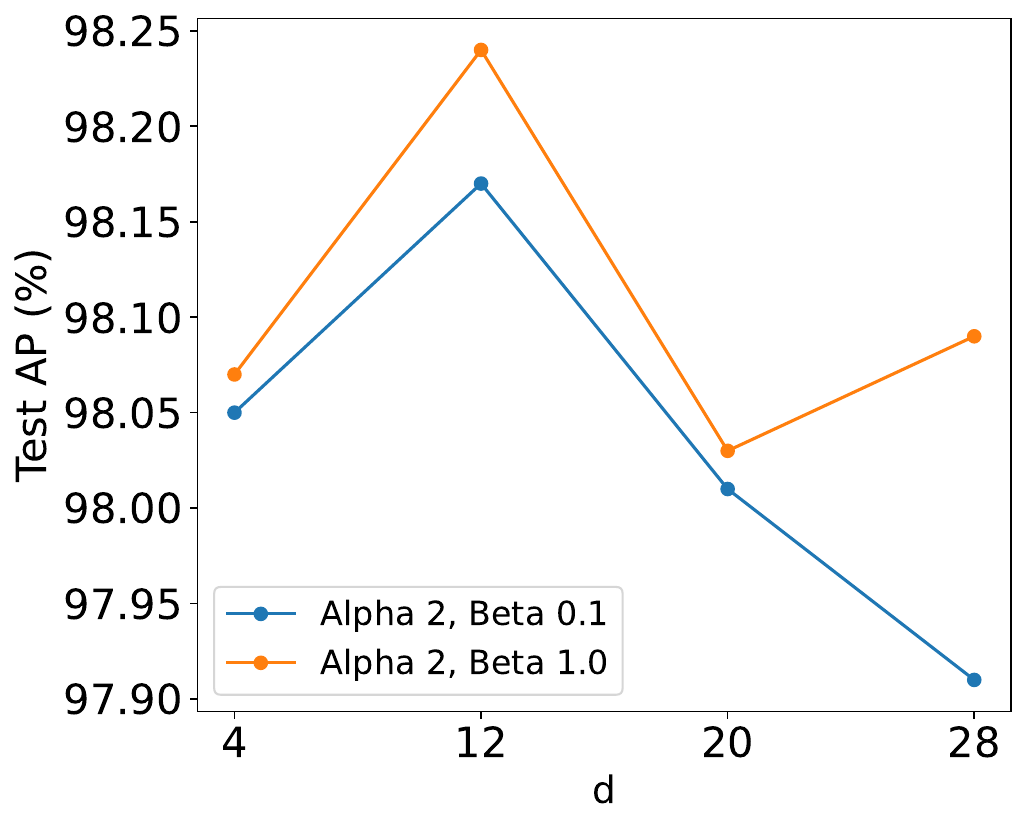}   
        \end{minipage}
    }
    \subfloat[Transductive on Wikipedia-fm] 
    {
        \begin{minipage}[t]{0.23\textwidth}
            \centering      
            \includegraphics[width=1\textwidth]{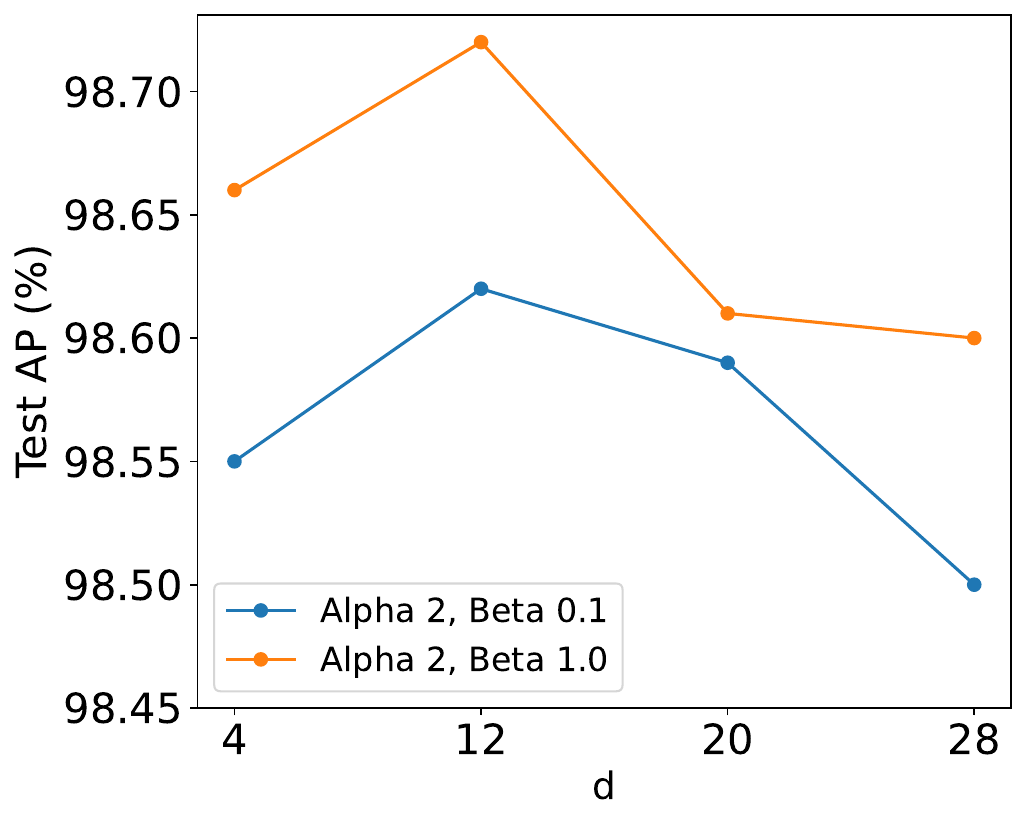}   
        \end{minipage}
    }
    \caption{Hyper-parameter analysis with effective \(\alpha = 2\).} 
    \label{fig:alpha2_analysis}  
\end{figure}

\begin{figure}[htbp]
    \centering    
    \subfloat[Inductive on Wikipedia]
    {
        \begin{minipage}[t]{0.23\textwidth}
            \centering    
            \includegraphics[width=1\textwidth]{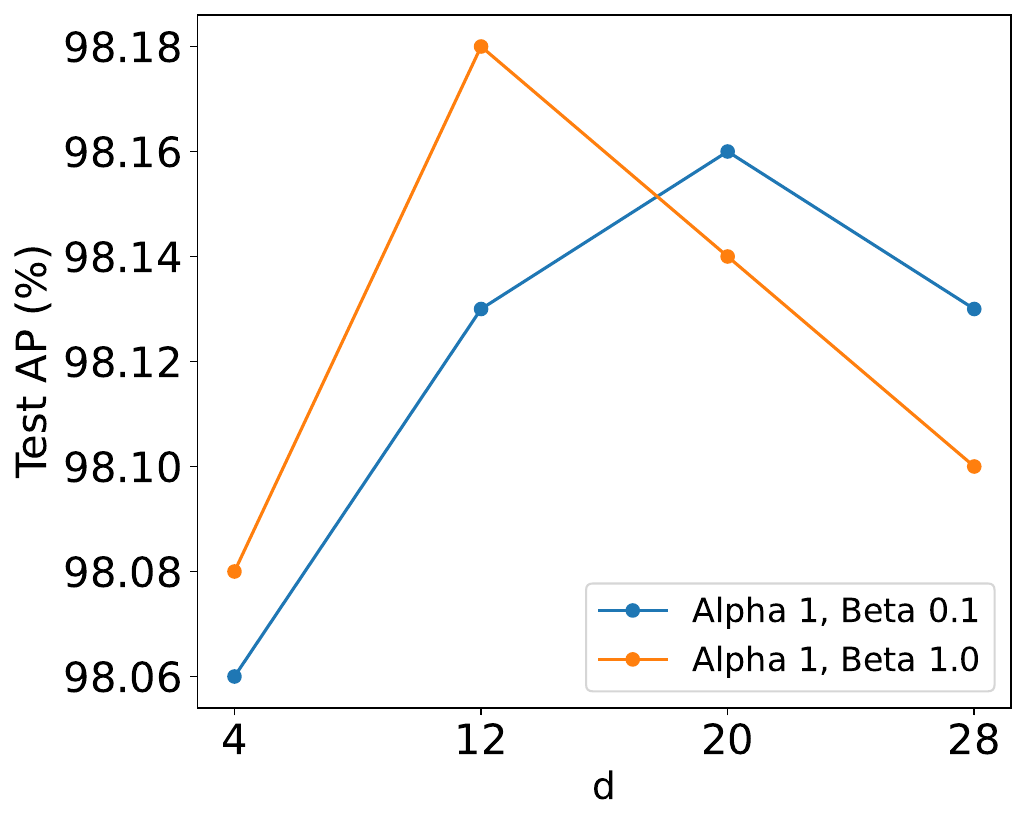}   
        \end{minipage}
    }
    \subfloat[Transductive on Wikipedia] 
    {
        \begin{minipage}[t]{0.23\textwidth}
            \centering    
            \includegraphics[width=1\textwidth]{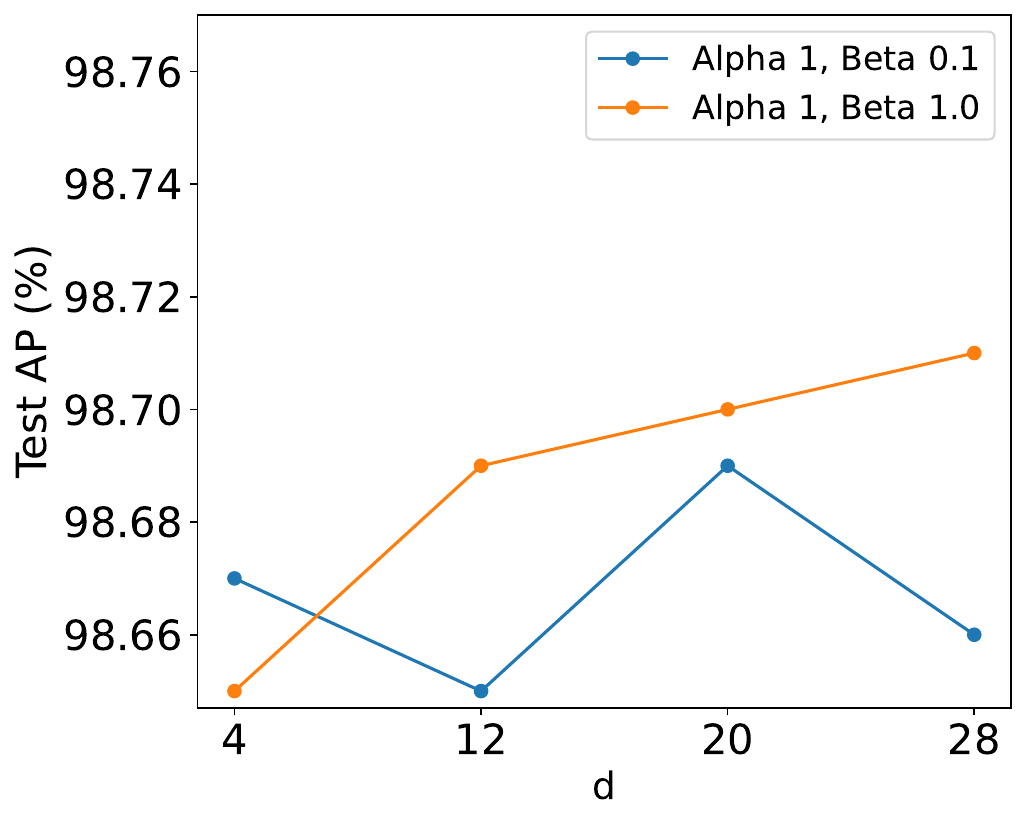}   
        \end{minipage}
    }
    
    \subfloat[Inductive on Wikipedia-fm] 
    {
        \begin{minipage}[t]{0.23\textwidth}
            \centering      
            \includegraphics[width=1\textwidth]{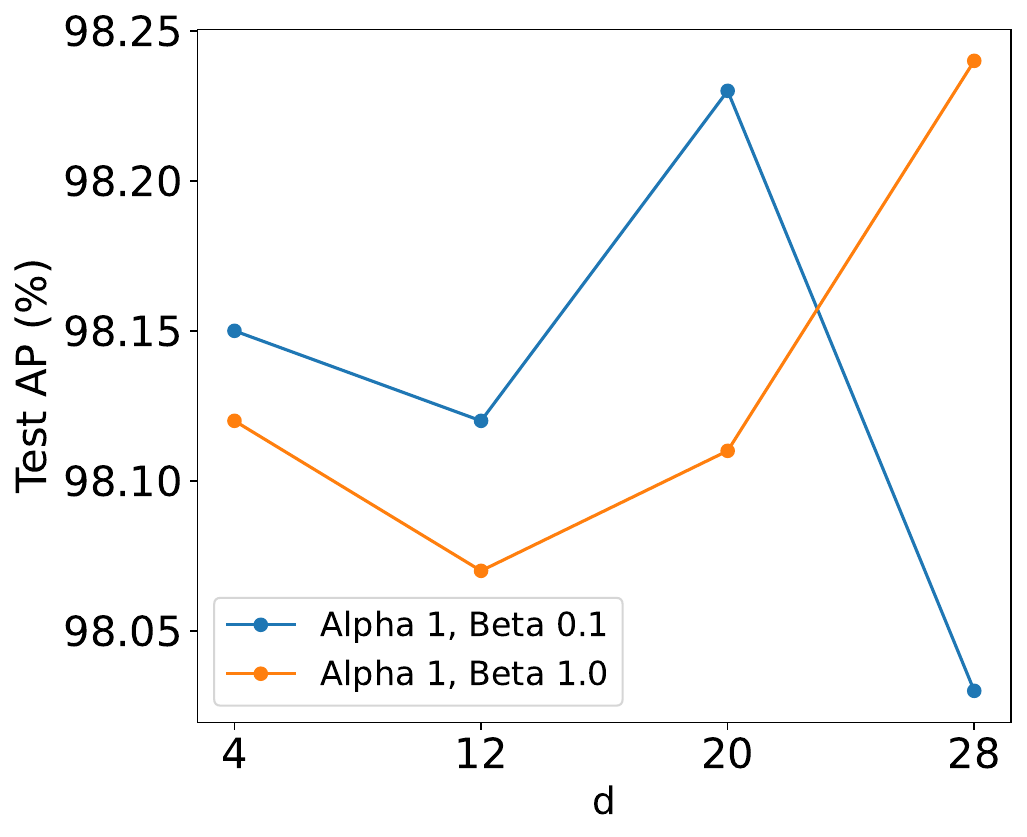}   
        \end{minipage}
    }
    \subfloat[Transductive on Wikipedia-fm] 
    {
        \begin{minipage}[t]{0.23\textwidth}
            \centering      
            \includegraphics[width=1\textwidth]{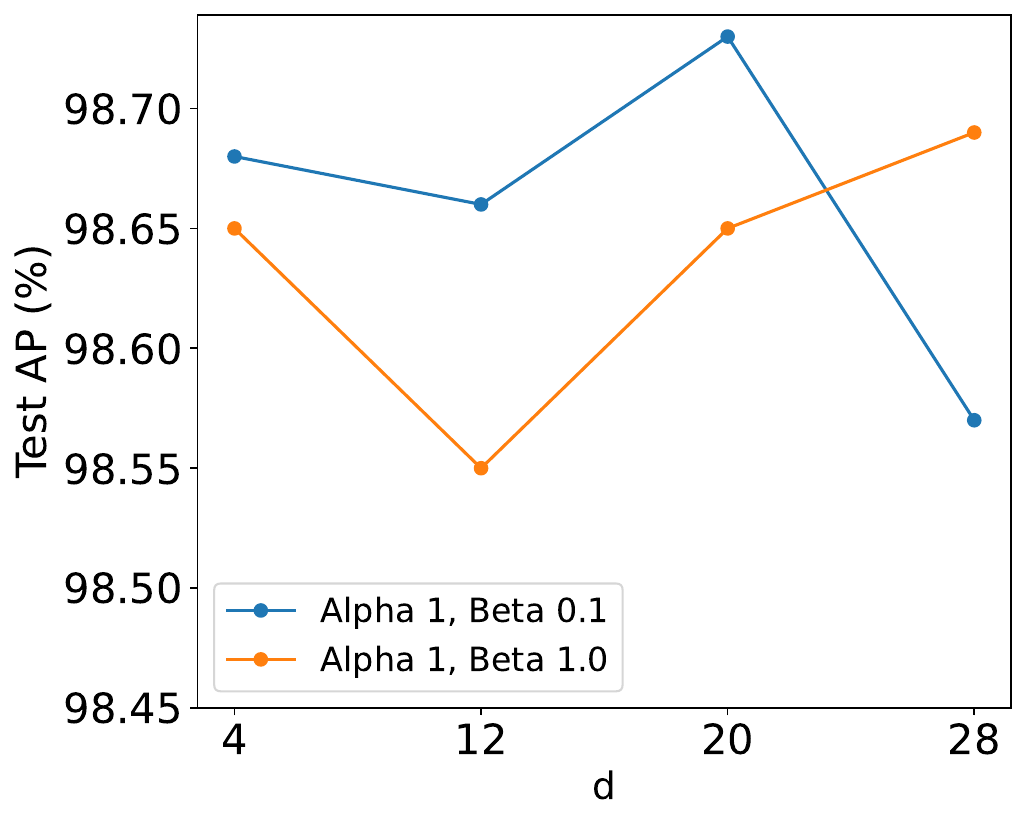}   
        \end{minipage}
    }
    \caption{Hyper-parameter analysis with ineffective \(\alpha = 1\)} 
    \label{fig:alpha1_analysis}  
\end{figure}

In this section, we investigate the impact of the scale parameter \(\alpha\) and the time decay factor \(\beta\) in both their ineffective and effective states, as well as the role of the dimension \(d\). The mean AP, averaged over 10 runs, is reported. Notably, when \(\alpha = 1\), the step count is ineffective (Figure~\ref{fig:alpha1_analysis}), while an effective \(\alpha = 2\) (Figure~\ref{fig:alpha2_analysis}) yields significant performance differences:
\begin{itemize}
    \item For the attributed dataset Wikipedia, the best performance is achieved in both induction and transduction tasks when the dimensionality \(d\) is relatively small, specifically \(d = 4\). Moreover, across different combinations of \(\alpha\) and \(\beta\), the overall performance tends to decline as \(d\) increases. This suggests that the model primarily relies on the dataset's inherent attributes, with trajectory encoding as a secondary role, allowing for a smaller dimensionality.
    \item For the non-attributed Wikipedia-fm dataset, whether in transductive or inductive tasks, the optimal dimensionality \(d\) increases across different combinations of \(\alpha\) and \(\beta\). This indicates that in the absence of inherent node and edge features, the model relies more heavily on trajectory encoding to capture the underlying temporal and structural information, necessitating a higher dimensionality to represent these relationships effectively.
    \item When \(\alpha\) is effective (\(\alpha = 2\)), it is observed that on the Wikipedia dataset, both transductive and inductive tasks consistently perform better when \(\beta = 0.1\) compared to \(\beta = 1\). Similarly, on the Wikipedia-fm dataset, performance is consistently better when \(\beta = 1\) compared to \(\beta = 0.1\). Notably, when \(\alpha\) is effective, the optimal dimensionality \(d\) is also consistent across tasks. 
    \item When \(\alpha\) is ineffective (\(\alpha = 1\)), the performance in transductive and inductive tasks does not consistently surpass other settings, and the peak performance shifts as \(d\) changes. Moreover, cases with effective \(\alpha = 2\) consistently outperform those with ineffective \(\alpha = 1\). This highlights the critical role of \(\alpha\) as a scaling factor in capturing temporal dependencies and its significant contribution to model robustness and effectiveness. The different optimal values of \(\beta\) across datasets suggest that \(\beta\) adjusts to the specific characteristics of each dataset.
\end{itemize}

\subsection{Node Classification}
\begin{table}[ht]
\centering
\caption{Comparison of node classification performance (AUC, mean $\pm$ std, averaged by 5 runs) on Wikipedia and Reddit datasets. The best results are marked in bold.}
\label{tab:node_classification}
\begin{tabular}{@{}lcc@{}}
\toprule
\textbf{Model} & \textbf{Wikipedia} & \textbf{Reddit} \\
\hline
JODIE    & 83.91 $\pm$ 0.3 & 61.93 $\pm$ 0.3\\
TGAT     & 83.71 $\pm$ 0.2 & 65.56 $\pm$ 0.5\\
DyRep    & 83.29 $\pm$ 0.3 & 62.61 $\pm$ 0.5\\
TGN-Att  & \underline{87.81} $\pm$ 0.6& 67.26 $\pm$ 0.6\\
PINT     & 87.04 $\pm$ 0.3& \underline{67.34} $\pm$ 0.5\\
\cmidrule(lr){1-3}
 TETGN (ours)& \textbf{88.21 $\pm$ 0.4}&\textbf{67.71 $\pm$ 0.3}\\
\bottomrule
\end{tabular}
\end{table}
% In Table~\ref{tab:node_classification}, we present the node classification results for Wikipedia and Reddit. Overall, \textbf{TETGN} achieves the best performance on both datasets. Compared to the strongest baselines, TETGN outperforms TGN-Att by approximately 0.4\% on Wikipedia and PINT by around 0.37\% on Reddit. Moreover, the relatively low standard deviations indicate that TETGN also maintains stable results. These gains highlight the effectiveness and versatility of our trajectory encoding module in different tasks.
In Table~\ref{tab:node_classification}, we compare the performance of several models on node classification tasks over the Wikipedia and Reddit datasets. We make the following observations:

\begin{itemize}
    \item \textbf{Superior Performance.} TETGN achieves the highest AUC scores on both Wikipedia and Reddit. This outperforms the strongest baselines by a margin of approximately 0.4\% on Wikipedia (compared to TGN-Att) and 0.37\% on Reddit (compared to PINT). 
    \item \textbf{Variance Stability.} The relatively low standard deviations highlight that TETGN not only improves average performance but also maintains consistent results across multiple runs. This stability suggests that our trajectory encoding mechanism yields better node embeddings in the face of temporal dynamics.
    % \item \textbf{Impact of Trajectory Encoding.} The improvements can be largely attributed to our trajectory encoding module, which captures both the historical context and relative positional information of each node. This design allows the model to better capture subtle structural differences and temporal dependencies that traditional TGNs may overlook.

\end{itemize}
% It shows that TETGN has versatility across various dynamic graph tasks.
These results show that TETGN is effective and versatile for different downstream tasks.

\section{Conclusion, Limitation and Future Work}\label{sec:conclusion}

This paper introduced TETGN to address longstanding limitations in both anonymous and non-anonymous Temporal Graph Networks (TGNs). By encoding node trajectories within the message-passing process, TETGN effectively captures temporal and positional relationships without requiring complex preliminary walks. As a result, it can distinguish nodes in symmetric or cyclic structures where previous methods fall short. Experimental results on three real-world datasets confirm its superior performance in modelling dynamic graphs. 

Whilst TETGN demonstrates strong performance, we acknowledge two limitations: (1) The dual message-passing streams introduce additional computational overhead; (2) The exponential trajectory encoding may miss periodic patterns in temporal interactions. Future work could explore more efficient encoding mechanisms and extend to large-scale complex temporal networks.\

Overall, TETGN takes a step forward in unifying anonymous and non-anonymous approaches, and we anticipate it will inspire further innovations in temporal graph representation learning.

\bibliographystyle{IEEEtran}
% \bibliography{references}
\input{main.bbl}

\end{document}